\documentclass[runningheads]{llncs}

\usepackage{amsmath,amsfonts}
\usepackage{graphicx}
\usepackage{booktabs}
\usepackage{multirow}
\usepackage{caption}
\usepackage{float}
\usepackage{tabularx}
\usepackage{comment}
\usepackage{siunitx}
\newcolumntype{Y}{>{\raggedleft\arraybackslash}X} 
\newcolumntype{L}{>{\raggedright\arraybackslash}X}
\usepackage{algorithm, algpseudocode}

\title{Machine Learning Predictions for Traffic Equilibria in Road Renovation Scheduling}

\author{Robbert Bosch \orcidID{0009-0006-8807-1031} \and Wouter van Heeswijk \orcidID{0000-0002-5413-9660} \and Patricia Rogetzer \orcidID{0000-0001-9582-6320} \and Martijn Mes \orcidID{0000-0001-9676-5259}}
\institute{University of Twente, Department of High Tech Business and Entrepeneurship, The Netherlands \\\email{\{r.r.bosch, w.j.a.vanheeswijk, p.b.rogetzer, m.r.k.mes\}@utwente.nl}}
\titlerunning{Machine Learning Predictions for Traffic Equilibria} 
\begin{document}

\begin{abstract}
Accurately estimating the impact of road maintenance schedules on traffic conditions is important because maintenance operations can substantially worsen congestion if not carefully planned. Reliable estimates allow planners to avoid excessive delays during periods of roadwork. Since the exact increase in congestion is difficult to predict analytically, traffic simulations are commonly used to assess the redistribution of the flow of traffic. However, when applied to long-term maintenance planning involving many overlapping projects and scheduling alternatives, these simulations must be run thousands of times, resulting in a significant computational burden. This paper investigates the use of machine learning-based surrogate models to predict network-wide congestion caused by simultaneous road renovations. We frame the problem as a supervised learning task, using one-hot encodings, engineered traffic features, and heuristic approximations. A range of linear, ensemble-based, probabilistic, and neural regression models is evaluated under an online learning framework in which data progressively becomes available. The experimental results show that the Costliest Subset Heuristic provides a reasonable approximation when limited training data is available, and that most regression models fail to outperform it, with the exception of XGBoost, which achieves substantially better accuracy. In overall performance, XGBoost significantly outperforms alternatives in a range of metrics, most strikingly Mean Absolute Percentage Error (MAPE) and Pinball loss, where it achieves a MAPE of 11\% and outperforms the next-best model by 20\% and 38\% respectively. This modeling approach has the potential to reduce the computational burden of large-scale traffic assignment problems in maintenance planning.
\end{abstract}


\maketitle

\section{Introduction}
\label{sec:introduction}
This study is motivated by the need to efficiently evaluate maintenance schedules by predicting the traffic impact of simultaneous road renovations. The inspiration for this work is the Road Network Maintenance Scheduling Problem (RNMSP), which involves determining the optimal timing of maintenance activities on a road network to minimize disruption while respecting budgetary, resource, and deadline constraints. We specifically address the computational burden of traffic simulations through surrogate modeling, focusing on a fixed instance in which a predefined set of roads must be renovated. Due to the urgency of many renovation tasks, multiple projects must be scheduled simultaneously, with the objective of minimizing overall congestion while adhering to project deadlines.

\begin{figure}
    \centering
    \includegraphics[width=0.5\linewidth]{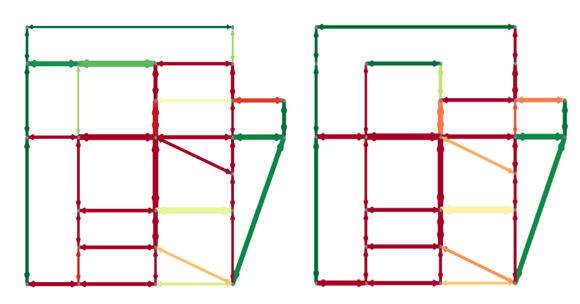}
    \caption{Illustration of the Sioux Falls traffic network. Left shows the graph of the traffic network in a normal situation, right shows the network when two roads are removed.}
    \label{fig:SF_changes}
\end{figure}

As a case study, we use the Sioux Falls traffic network, which represents a simplified highway system with predefined traffic demand. The network is modeled as a directed graph, where each link represents a road with a given free flow travel time ($FFT$) and capacity. Renovation of a road temporarily reduces its capacity and increases its $FFT$, causing traffic to reroute and potentially increasing congestion throughout the network. Figure~\ref{fig:SF_changes} shows the Sioux Falls network, with uncongested traffic flows in green and congested flows in red. The right network illustrates the effect of removing two roads, redistributing traffic.

Each schedule specifies which maintenance projects are active in each time period. Each renovation project is linked to a road, which is closed for its duration. Each unique combination of closed roads, hereafter called a scenario, requires a traffic simulation to calculate the total travel time ($TTT$), defined as the aggregated travel time for all vehicles on the network. The cumulative increase in $TTT$ across all time periods is used as a performance measure for the schedule. 

Traffic simulations to estimate the traffic impact of renovations are commonly executed using a macroscopic traffic model, called a Traffic Assignment Problem (TAP). While accurate, these simulations are computationally intensive, with run times ranging from seconds for small networks to minutes for larger ones. Optimizing a schedule often requires evaluating thousands of alternatives, each involving multiple TAP simulations, making the total computational cost prohibitive.

To reduce the computational burden of traffic simulations, we investigate surrogate models that estimate the $TTT$ outcomes of unevaluated scenarios, enabling early identification of suboptimal schedules. If a regression model predicts that a schedule is likely to lead to a high $TTT$, it can be discarded without simulating any further associated scenarios. This approach can significantly reduce the number of simulations required.

To be effective, a surrogate must have several characteristics. First, a surrogate must be data-efficient. In the early stages of the optimization process, only a small number of scenarios will have been simulated, which limits the amount of training data available. A surrogate model that requires a large number of simulations to generalize well would offer limited benefit. As such, the models considered in this study must demonstrate reasonable predictive performance even when trained on a limited number of observations.

Second, the surrogate model must be suitable for use in an online optimization setting, in which new scenarios are evaluated incrementally. This requires the model either to be capable of online updates or to have sufficiently fast training times to be retrained periodically as new data becomes available. Surrogates that require extensive retraining or prohibitively long fitting times are impractical for use within the iterative structure of schedule optimization algorithms.

Third, to avoid discarding promising schedules, we prefer surrogate models that make conservative predictions, slightly underestimating congestion. This is done using several techniques: Quantile regression, Bayesian regression, and ensemble methods. The amount of conservatism can be tuned via parameters to the desired level depending on the technique.

The remainder of the paper is structured as follows. In Section \ref{sec:related_work} we discuss papers related to the RNMSP and surrogate models for the TAP. In Section~\ref{sec:problem_description} we discuss the definition of the TAP and the regression-based estimation task.
In Section \ref{sec:methodology} we discuss our proposed methodology, which we test in Section \ref{sec:experiments}. Section \ref{sec:conclusion} concludes this paper.

\section{Related Work}
\label{sec:related_work}

The Road Network Maintenance Scheduling Problem (RNMSP) has been the subject of considerable research interest due to its relevance in minimizing the congestion caused by planned maintenance on urban and regional traffic networks. This work is related to three primary areas of the literature: (i) road work scheduling problems (RWSPs), particularly RNMSPs; (ii) traffic simulation and surrogate modeling; and (iii) regression-based approaches for efficient evaluation within large-scale combinatorial optimization.

The RNMSP considers scheduling road maintenance or rehabilitation projects across a network while accounting for their impact on congestion. Previous research has focused on optimizing scheduling under various objectives such as minimizing $TTT$, cost, or environmental impact~\cite{cheu_genetic_2004,gong_optimizing_2016,zheng_measuring_2014,song_rehabilitation_2018}. Population-based metaheuristics, such as genetic algorithms (GA), are often used for this purpose due to their ability to handle complex, multi-objective, and black-box optimization problems~\cite{miralinaghi_contract_2022,aksoy_urban_2021,jiang_time-dependent_2015}. However, as the RNMSP is formulated as a bi-level optimization problem, where schedules are generated at the top level and traffic assignments are generated at the bottom level, this requires solving many traffic simulations for each schedule configuration. This makes these solution methods computationally intractable as problem size increases.

Traffic simulations, which estimate the impact of road closures on network-wide flow, are frequently implemented at the macroscopic level using user equilibrium or system-optimal assignment models~\cite{chen_roadside_2024,li_bi-level_2021}. However, due to the amount of traffic simulations that need to be evaluated in the optimization process, the computational burden of these algorithms is very large. This challenge is exacerbated not only by large networks, but more critically by the large number of renovation schedules that must be evaluated. To reduce this burden, some studies have explored simplified models or heuristics~\cite{kumar_simplified_2018,lu_optimal_2016}, although often at the cost of solution quality. 

Although surrogate modeling has been used in transportation to reduce the cost of repeated simulations, regression-based surrogate models remain relatively underexplored, particularly for predicting congestion in the context of road maintenance scheduling. Bagloee et al.~\cite{bagloee_optimization_2018} propose a hybrid framework that combines machine learning with linear programming to approximate the traffic impacts of road construction projects. Gu et al.~\cite{gu_surrogate-based_2019} apply surrogate modeling for toll optimization, demonstrating how learned models can accelerate decisions in congested networks. More recently, Natterer et al.~\cite{natterer_machine_2025} use graph neural networks to approximate agent-based traffic simulations, showing promising results for network-scale surrogate prediction. However, few studies systematically explore how different regression models and feature sets perform under limited data availability or under requirements for conservative predictions. This motivates our comprehensive evaluation of regression models and feature encodings for surrogate-based congestion estimation.

\textbf{Contribution.} This paper contributes to the RNMSP literature and related fields that require repeated traffic simulations by evaluating a broad range of regression models for predicting the total travel time caused by road closures. We explore multiple feature representations, including one-hot encodings of renovation projects and engineered features derived from baseline traffic data. In contrast to existing approaches, we systematically evaluate model performance under limited data availability and investigate models' ability to produce predictions that slightly underestimate congestion in order to reduce the risk of discarding promising schedules. These contributions enable more scalable and efficient surrogate-assisted optimization for large-scale RNMSPs.

\section{Problem Description}
\label{sec:problem_description}
In this section, we define the problem of estimating the $TTT$ in a traffic network under road renovation scenarios. The primary computational challenge lies in the repeated evaluation of $TTT$ through traffic simulation, which requires solving the TAP, an iterative and computationally expensive procedure. As evaluating a schedule requires solving many TAPs, this becomes prohibitively expensive to compute. As a solution, we propose the use of supervised regression models to approximate the outcome of the TAP when given a set of closed roads in a fixed network.

\subsection{Traffic Assignment Problem (TAP)}
\label{subsec:tap}

The TAP models the equilibrium distribution of traffic across a network. Let $\mathcal{G} = (\mathcal{N}, \mathcal{A})$ be a directed graph representing the traffic network, where $\mathcal{N}$ denotes the set of nodes and $\mathcal{A}$ the set of directed links. Traffic demand is specified by a set of origin-destination (O-D) pairs $\mathcal{W} \subseteq \mathcal{N} \times \mathcal{N}$, where each $w \in \mathcal{W}$ has a fixed demand $m_w$. For each O-D pair $w$, traffic may be assigned to any path $p \in \Pi_w$ that connects the origin to the destination.

When a subset of roads $\mathcal{A'} \subseteq \mathcal{A}$ is closed for renovation, the operational network becomes $\mathcal{A} \setminus\mathcal{A}'$. The closure reduces the effective capacity and increases the free-flow travel time on affected links, altering the cost of travel and thus the resulting equilibrium flows. The TAP determines the new assignment of flows, ensuring that no road user can reduce their travel time by unilaterally changing routes. The total travel time $TTT(\mathcal{A}')$ under closure configuration $\mathcal{A}'$ is defined as the sum of all travel times experienced by vehicles in the network. Table~\ref{tab:notation_tap} summarizes the main notation used in this formulation.
\begin{table}[h]
\centering
\caption{Notation for the Traffic Assignment Problem and renovation scenarios}
\label{tab:notation_tap}
\begin{tabularx}{\textwidth}{lX}
\toprule
\textbf{Symbol} & \textbf{Description} \\
\midrule
$\mathcal{G} = (\mathcal{N}, \mathcal{A})$ & Directed traffic network with nodes $\mathcal{N}$ and links $\mathcal{A}$ \\
$(i,j) \in \mathcal{A}$ & Directed link between nodes $i$ and $j$ \\
$W \subseteq \mathcal{N} \times \mathcal{N}$ & Set of origin-destination (OD) pairs \\
$\Pi_w$ & Set of feasible paths for OD pair $w \in W$ \\
$\mathcal{A}' \subseteq \mathcal{A}$ & Set of roads closed in a scenario \\
$m_w$ & Travel demand for OD pair $w$ \\
$r_{ij}^{\text{base}}$ & Free-flow travel time on link $(i,j)$ in the baseline network \\
$\sigma_{ij}^{\text{base}}$ & Capacity of link $(i,j)$ in the baseline \\
$\sigma_{ij}^{p}$ & Adjusted capacity of link $(i,j)$ under project $p$ \\
$r_{ij}^{p}$ & Adjusted free-flow time of link $(i,j)$ under project $p$ \\
$a$, $b$ & Parameters of the BPR function (typically $a = 0.15$, $b = 4$) \\
$G_{tp}$ & Indicator: 1 if project $p$ is active at time $t$, 0 otherwise \\
$x_p$ & Flow assigned to path $p \in \Pi_w$ \\
$f_{ij}$ & Total flow on link $(i,j)$ \\
$c_{ij}$ & Travel cost on link $(i,j)$ \\
$c_p$ & Travel cost of path $p$ \\
$\lambda_w$ & Minimal path cost for OD pair $w$ (used in equilibrium conditions) \\
$\delta_{ij}^p$ & Indicator: 1 if link $(i,j)$ is part of path $p$, 0 otherwise \\
$TTT(\mathcal{A}')$ & Total travel time under scenario $\mathcal{A}'$ \\
\bottomrule
\end{tabularx}
\end{table}

The TAP is commonly formulated as a convex optimization problem~\cite{beckmann_studies_1956}, where the objective is to minimize the total travel cost across all flows by assigning traffic flow along paths $x_p$:
\begin{equation}
\label{eq:UE_objective}
    \min_{x_p \in \Pi_w} \sum_{w \in \mathcal{W}} \sum_{p \in \Pi_w} \int_0^{x_p} c_p(z) \, dz
\end{equation}
subject to:
\[
\sum_{p \in \Pi_w} x_p = m_w, \quad x_p \geq 0 \quad \forall w \in \mathcal{W}, \forall p \in \Pi_w
\]
\[
c_p - \lambda_w \geq 0, \quad (c_p - \lambda_w) x_p = 0 \quad \forall p \in \Pi_w
\]

Path costs are computed as the sum of link costs:
\begin{equation}
\label{eq:path cost}
    c_p = \sum_{(i,j) \in \mathcal{A}} \delta_{ij}^p \cdot c_{ij}
\end{equation}
where $\delta_{ij}^p = 1$ if link $(i,j)$ is used in path $p$ and 0 otherwise. Link travel costs $c_{ij}$ are computed using the Bureau of Public Roads (BPR) function~\cite{roads_traffic_1964}:
\begin{equation}
\label{eq:link cost}
    c_{ij} = r_{ij} \left(1 + a \left(\frac{f_{ij}}{\sigma_{ij}} \right)^b \right), \quad \text{typically } a=0.15,\, b=4
\end{equation}

When a project is ongoing at time $t$, the capacity and free-flow time of affected links are adjusted as follows:
\begin{equation}
\label{eq:link capacity adjustment}
    \sigma_{ij}^{t} =
    \begin{cases}
        \sigma_{ij}^{\text{base}}, & \text{if } G_{tp}=0 \\
        \sigma_{ij}^{p}, & \text{if } G_{tp}=1
    \end{cases}, \quad \forall t, \forall (i,j) \in \mathcal{A}
\end{equation}
\begin{equation}
\label{eq:link fft adjustment}
    r_{ij}^{t} =
    \begin{cases}
        r_{ij}^{\text{base}}, & \text{if } G_{tp}=0 \\
        r_{ij}^{p}, & \text{if } G_{tp}=1
    \end{cases}, \quad \forall t, \forall (i,j) \in \mathcal{A}
\end{equation}

\subsection{TAP Solution Method}
\label{subsec:frank_wolfe}

To solve the TAP, we employ the Frank-Wolfe algorithm~\cite{nguyen_algorithm_1974}, the most widely adopted method to calculate the user equilibrium in traffic networks. The algorithm iteratively updates flows by solving shortest-path problems based on current link costs and combining solutions via a line search. Although the Frank-Wolfe algorithm is computationally efficient because it exploits the convex structure of the TAP, it remains an iterative procedure that requires repeatedly solving shortest-path problems across all origin-destination pairs, which can be computationally intensive for large networks or when many scenarios must be evaluated. The procedure is outlined in Algorithm~\ref{algo:TAP}. The resulting link and path flows define the equilibrium state, from which the total travel time \(TTT(\mathcal{A}')\) is calculated.

\begin{algorithm}
    \caption{Traffic Assignment Algorithm (Frank-Wolfe)}
    \label{algo:TAP}
    \begin{algorithmic}[1]
        \State \textbf{Initialization:} Assign all demand to shortest paths based on free-flow times \(r_{ij}\). Set \(k = 0\).
        \While{not converged}
            \State Update link costs \(c_{ij}^{k}\) using the BPR function.
            \State Compute shortest paths for each OD pair using \(c_{ij}^{k}\).
            \State Perform all-or-nothing assignment to obtain auxiliary flows \(y_{ij}^{k}\).
            \State Compute step size \(\lambda^{k}\) via line search to minimize total cost.
            \State Update flows: \(f_{ij}^{k+1} = \lambda^{k} y_{ij}^{k} + (1 - \lambda^{k}) f_{ij}^{k}\).
            \State Check for convergence; if not satisfied, increment \(k \leftarrow k + 1\) and repeat.
        \EndWhile
    \end{algorithmic}
\end{algorithm}

\subsection{Regression-Based Estimation Task}
\label{subsec:regression_task}

The regression-based estimation task defines a supervised learning problem in which the goal is to predict the total travel time \( TTT(\mathcal{A}') \) directly from the properties of a closure configuration \(\mathcal{A}' \subseteq \mathcal{A} \), bypassing the need to explicitly solve the Traffic Assignment Problem. In Table~\ref{tab:notation_regression} we provide the notation of the estimation task. Each data point consists of a closure configuration \(\mathcal{A}'_i \) and an associated target value \( y_i = TTT(\mathcal{A}'_i)\). The target is obtained by solving the TAP for \(\mathcal{A}'_i \) to compute \( TTT(\mathcal{A}'_i) \). The surrogate model is trained on \( M \) closure-target pairs \( (\mathcal{A}'_i, y_i) \). To encode the input configuration $\mathcal{A}'$ we construct features that capture both structural and flow-related properties of closed roads. The full set of representations is described in Section~\ref{subsec:feature_engineering}.

\begin{table}[h]
\caption{Main notation for the Regression-Based Estimation Task}
\label{tab:notation_regression}
\centering
\begin{tabularx}{\textwidth}{lX}
\toprule
\textbf{Symbol} & \textbf{Description} \\
\midrule
$\mathcal{A}' \subseteq \mathcal{A}$ & Set of closed roads (closure configuration) \\
$TTT(\mathcal{A}')$ & Total travel time from traffic simulation \\
$y = TTT(\mathcal{A}')$ & regression target \\
$M$ & Number of available data points \\
$\hat{y}$ & Predicted TTT \\
$\hat{f}$ & Regression model predicting \(\hat{y}\) given closure features \\
$X$ & Feature representation of \(\mathcal{A}'\) (one-hot, pairwise, engineered) \\
\bottomrule
\end{tabularx}
\end{table}

Because the surrogate model is used to eliminate poor-performing schedules before simulation, slight underestimation of congestion is preferred to overestimation. Conservative predictions help avoid prematurely discarding configurations that may in fact be viable. To achieve this, we consider several strategies which we explain further in Section~\ref{subsec:regression_models}.

The regression-based estimation task defines a supervised learning problem in which the goal is to predict \( TTT(\mathcal{A}') \) using limited training data, balancing predictive accuracy with a controlled conservative bias. The surrogate model should achieve sufficiently accurate predictions while favoring slight underestimations to support effective pruning during optimization. Section \ref{sec:methodology} describes the feature engineering, model selection in detail.

\section{Methodology}
\label{sec:methodology}

This section presents the methodology used to predict the total travel time ($TTT$) for road closure configurations using supervised learning. We describe the construction of the input features, followed by the feature selection process, and finally detail the regression models and methods for promoting conservative predictions. Experimental results are deferred to Section~\ref{sec:experiments}.

\subsection{Feature Engineering}
\label{subsec:feature_engineering}

Each training example corresponds to a subset $\mathcal{A}' \subseteq \mathcal{A}$ of roads closed due to renovation. We consider three types of input representations. The first is one-hot encoding, where each road in the network is represented as a binary vector of length equal to the total number of projects. Each index in this vector is set to 1 if the corresponding road is closed in $\mathcal{A}'$ and 0 otherwise. The second is pairwise encoding, which extends the one-hot representation by including binary features for each unique pair of roads (i.e. a 1 if  both roads are closed in $\mathcal{A'}$), capturing pairwise interactions between projects. 

The third representation involves engineered features, a compact set of metrics derived from domain knowledge and baseline network properties. These features include the project set size (the number of closed roads), centrality measures such as the total and average betweenness and closeness centrality of the closed roads, the naive impact (which reflects the cost contribution of closed roads in the baseline), and several statistical descriptors, including the sum, maximum, and mean of road-level properties in subset $\mathcal{A}'$.

The three feature representations are designed to capture both the topological characteristics and flow-related impacts of the closure set, providing the necessary information for the regression model to learn $TTT$ behavior across diverse renovation configurations.

\subsection{Feature Selection}
\label{subsec:feature_selection}

To improve reduce redundancy, we apply feature selection to the engineered features, beginning with a Pearson correlation analysis to assess the relationship between each feature and the target value.  Most features show correlations lower than 0.1 with $TTT$. However, some features such as Naive Impact, Sum of Disrupted Flow, and Project Set Size, achieve Pearson correlation coefficients above 0.6 with respect to $\log(TTT)$, indicating their predictive relevance and the usefulness of performing a logarithmic transformation on the prediction target. We then apply forward and backward sequential feature selection (FSFS and BSFS), to iteratively build feature subsets that maximize the validation $R^2$ score. The cut-off for feature selection is determined at the point where the marginal improvement in $R^2$ diminishes significantly, resulting in the selection of the nine best features for each method. The final feature set is the union of those selected by FSFS and BSFS, totaling 12 features. These selected features are used to train the regression models.

\subsection{Proposed Heuristics} 
\label{subsec:proposed_heuristics}

We propose three heuristic approaches to complement regression-based prediction models. These heuristics leverage the empirical observation that congestion generally increases as more roads are closed. Although this monotonicity does not hold universally due to Braess' paradox~\cite{braess_paradox_2005}, an analysis of 200{,}000 datapoints in this study showed no cases where closing additional roads reduced $TTT$.

The first heuristic, Costliest Subset Heuristic (CSH), uses the idea that for a set of closed roads \( \mathcal{A'} \), any subset \( \mathcal{A}'' \subseteq \mathcal{A}' \) whose $TTT$ is known can provide a soft lower bound for \( \mathcal{A} '\). Given a dataset of known configurations \( \mathcal{D} \), the heuristic searches all \( \mathcal{A}'' \in \mathcal{D} \) for which \( \mathcal{A}'' \subseteq \mathcal{A}' \), selecting the one with the highest observed $TTT$, see Equation~\ref{eq:csh}. If no subset is found, the $TTT$ associated with the empty set is returned. This approach is computationally efficient and provides a conservative estimate when sufficient subset evaluations are available.

\begin{equation}
\label{eq:csh}
\widehat{y}_{\text{CSH}}(\mathcal{A}') = \max_{\mathcal{A}'' \subseteq \mathcal{A}',\, \mathcal{A}'' \in \mathcal{D}} y(\mathcal{A}'')
\end{equation}
The second heuristic, Costliest Additive Subset (CASH), extends CSH by accounting for roads not included in the costliest subset. After identifying the subset of roads $\mathcal{A}^*$ that formed the result of $\widehat{y}_{\text{CSH}}(\mathcal{A}')$, the heuristic estimates the impact of the remaining roads \( \tilde{\mathcal{A}} = \mathcal{A}' \setminus \mathcal{A}^* \) using the same procedure, and adds the marginal increase, see Equation~\ref{eq:cash}. This method is more computationally intensive but may more accurately capture the total impact of road closures.
\begin{equation}
\label{eq:cash}
\widehat{y}_{\text{CASH}}(\mathcal{A}') = \widehat{y}_{\text{CSH}}(\mathcal{A}^*) + \widehat{y}_{\text{CSH}}(\tilde{\mathcal{A}}) - y(\emptyset)
\end{equation}
The third heuristic, Cheapest Superset Heuristic (CSupH), provides an upper bound by reversing the logic of CSH. It searches for supersets of the test configuration and selects the one with the lowest known $TTT$, shown in Equation~\ref{eq:csuph}:
\begin{equation}
\label{eq:csuph}
\widehat{y}_{\text{CSupH}}(\mathcal{A}') = \min_{\mathcal{A}'' \supseteq \mathcal{A}',\, \mathcal{A}' \in \mathcal{D}} y(\mathcal{A}'')
\end{equation}
This approach is useful for bracketing true outcomes between upper and lower estimates. The performance of these heuristics is evaluated in a simulated online learning environment, with results discussed in Section~\ref{subsec:heuristic_performance}.

\subsection{Proposed Regression Models}
\label{subsec:regression_models}
This subsection introduces the regression models evaluated in this study, specifying the feature representations used for training and the strategies applied to encourage conservative predictions under online learning conditions. 

For the main set of experiments, each closure configuration is described by a combination of features, including the one-hot vector indicating closed roads, the set of features selected in Section~\ref{subsec:feature_selection}, and the output of the heuristics described in Section~\ref{subsec:proposed_heuristics}. This expanded feature set aims to capture both the direct and indirect effects of road closures on the network. 

Conservative predictions are required to enable pruning during schedule optimization. We use three strategies to achieve this:

\begin{itemize}
    \item \textbf{Pinball loss functions:} Models are trained using pinball loss, which penalizes under- and over-estimation asymmetrically depending on a chosen quantile \( \tau \). For example, with $\tau=0.05$, the model will overestimate in 5\% of cases assuming the target variable is normal i.i.d. distributed. This approach is used in QuantileRegression, XGBoost, and the Neural Network.
    
    \item \textbf{Probabilistic regression:} Bayesian Ridge Regression estimates a posterior distribution over model parameters, which enables uncertainty-aware predictions. In our setup, the lower quantile of the predictive distribution is used as a conservative estimate.
    
    \item \textbf{Quantile prediction from ensembles:} For ensemble-based models such as Random Forest, Bagging Ridge, and \(k\)-Nearest Neighbors, we aggregate predictions from individual ensemble members and compute the empirical lower quantile. This provides a direct mechanism for conservative prediction using model diversity.
\end{itemize}

The models evaluated in this study are summarized in Table~\ref{tab:regression_models}. For all models with the ``Log'' prefix, the target variable is transformed using the natural logarithm prior to training to reduce the skewness of the $TTT$ distribution. Models without the ``Log'' prefix are trained directly on the original $TTT$ values. Full experimental results and model comparisons are presented in Section~\ref{sec:experiments}.
\begin{table}[h]
\centering
\caption{Regression models evaluated in this study}
\label{tab:regression_models}
\begin{tabularx}{\textwidth}{L L L L}
\toprule
\textbf{Name} & \textbf{Description} & \textbf{Target Variable} & \textbf{Conservative ~Prediction Method} \\
\midrule
LogOLS & Ordinary Least Squares & $\log(TTT)$ & None\\
LogQuantileReg & Linear Regression with Pinball loss & $\log(TTT)$ & Pinball loss\\
LogBayesianRidge & Bayesian Ridge regression & $\log(TTT)$ & Probabilistic regression\\
LogBagging & Bagging ensemble of OLS regressors & $\log(TTT)$ & Ensemble quantile\\
LogkNN & \(k\)-Nearest Neighbors regression & $\log(TTT)$ & Ensemble quantile\\
RF & Random Forest regression & $TTT$ & Ensemble quantile\\
XGBoost & Gradient Boosted Trees & $TTT$ & Pinball loss\\
NN & Neural Network & $TTT$ & Pinball loss\\
\bottomrule
\end{tabularx}
\end{table}

\section{Experimental setup and Results}
\label{sec:experiments}

This section presents the experimental evaluation of the regression models and feature representations described in Section~\ref{subsec:regression_models}. We aim to assess each model's predictive accuracy, conservativeness (i.e., tendency to underestimate $TTT$), and robustness to small dataset sizes.

\subsection{Experimental Setup} 
\label{subsec:experimental_setup}

The experimental validation is conducted on the Sioux Falls traffic network, a commonly used benchmark instance in transportation research~\cite{leblanc_efficient_1975}. The evaluation procedure simulates an online learning environment in which data becomes progressively available. Initially, no training data is provided. At each iteration, 1000 new datapoints are added to the training set. After training on all currently available data, the predictive models are evaluated on the next batch of 1000 unseen datapoints. In total, 200 iterations are performed unless computational limits are reached, for a total of 200,000 datapoints. To prevent infeasibly long evaluation times, a stopping criterion is imposed: if the median training and testing time per iteration over the last 10 iterations exceeds 600 seconds, the experiment is terminated prematurely. We summarize the metrics used to evaluate models in Table~\ref{tab:performance_metrics}.

\begin{table}[h]
\centering
\caption{Performance Metrics Used for Model Evaluation}
\label{tab:performance_metrics}
\begin{tabular}{lp{10cm}}
\toprule
\textbf{Metric} & \textbf{Description} \\
\midrule
MAE & Mean Absolute Error, measuring the average absolute difference between predicted and actual values. \\
Pinball Loss & Quantile loss function scaled by a factor depending on the quantile $\tau$: \\[0.3cm]
& $
L_{\tau}(y, \hat{y}) =
\begin{cases}
\tau (y - \hat{y}), & \text{if } y \geq \hat{y} \\
(1 - \tau)(\hat{y} - y), & \text{if } y < \hat{y}
\end{cases}
$ \\[0.3cm]
Bias & Mean Error, representing the average difference between predicted and actual values. \\
MAPE & Mean Absolute Percentage Error, indicating the average percentage difference between predictions and actual values. \\
Time & The combined training and prediction time per iteration, expressed in seconds. \\
\bottomrule
\end{tabular}
\end{table}

\subsection{Heuristic Performance} 
\label{subsec:heuristic_performance}

The three proposed heuristics are first evaluated using the experimental procedure described in Section~\ref{subsec:experimental_setup}. The results, summarized in the first three rows of  Table~\ref{tab:regressor_results}, indicate substantial differences in predictive performance.

Among the heuristics, the Costliest Subset Heuristic (CSH) achieves the best performance, both in terms of general predictive accuracy and in terms of conservative accuracy as measured by the pinball loss. In contrast, the AdditiveSubset and CheapestSuperset heuristics perform substantially worse. Based on these results, only the CSH heuristic is retained as an auxiliary feature for subsequent regression model training.

\subsection{Regression Model Performance} 
\label{subsec:regression_performance}

Following the heuristic evaluation, the regression models described in Section~\ref{subsec:experimental_setup} are trained and evaluated using the same online experimental design. Average performance results across all iterations are reported in Table~\ref{tab:regressor_results}.
\begin{table}[h]
\centering
\caption{Average heuristic and regression model performance metrics over iterations. *: Model exceeded allowed training time at the iteration denoted by *. **: Models had outliers during testing which were removed for averages.}
\label{tab:regressor_results}
\begin{tabularx}{\textwidth}{l Y Y Y Y Y}
\toprule
\textbf{Regressor} & \textbf{MAE ($\times 10^3$)} & \textbf{Pinball (0.05) ($\times 10^3$)} & \textbf{Bias ($\times 10^3$)} & \textbf{MAPE} & \textbf{Time/ Iteration (s)} \\
\midrule
AdditiveSubset & 11,968 & 6,658 & 1,498 & 49 & 6.4 \\
CostliestSubset & 11,408 & 570 & -11,408 & 30 & 5.2 \\
CheapestSuperset & 4,448,281 & 4,225,837 & 4,448,215 & 21,553 & 2.6 \\
LogBaggingRidge** & 8,917 & 2,751 & -3,795 & 24 & 20.0 \\
LogBayesianRidge** & 12,190 & 646 & -12,108 & 38 & 4.6 \\
LogOLS** & 9,197 & 3,365 & -2,740 & 26 & 6.5 \\
LogQuantileRegression & 12,319 & 920 & -11,643 & 33 & *:25 \\
LogkNN & 9,347 & 4,206 & -1,039 & 27 & 16.2 \\
NN & 10,440 & 572 & -10,329 & 26 & *:102 \\
RF & 21,095 & 17,915 & 16,372 & 66 & *:143 \\
XGBoost & 8,042 & 460 & -7,912 & 15 & 39.0 \\
\bottomrule
\end{tabularx}
\end{table}

Among the tested models, LogQuantileRegression, Neural Network, and Random Forest exceeded the computational time limit and did not complete all 200 iterations, instead being cut off at the iteration denoted by * in Table~\ref{tab:regressor_results}. Despite this, the Neural Network achieved strong predictive performance during the iterations it did complete, suggesting that it may have performed competitively if allowed to run to completion.

\begin{figure}[h]
    \centering
    \includegraphics[width=1\linewidth]{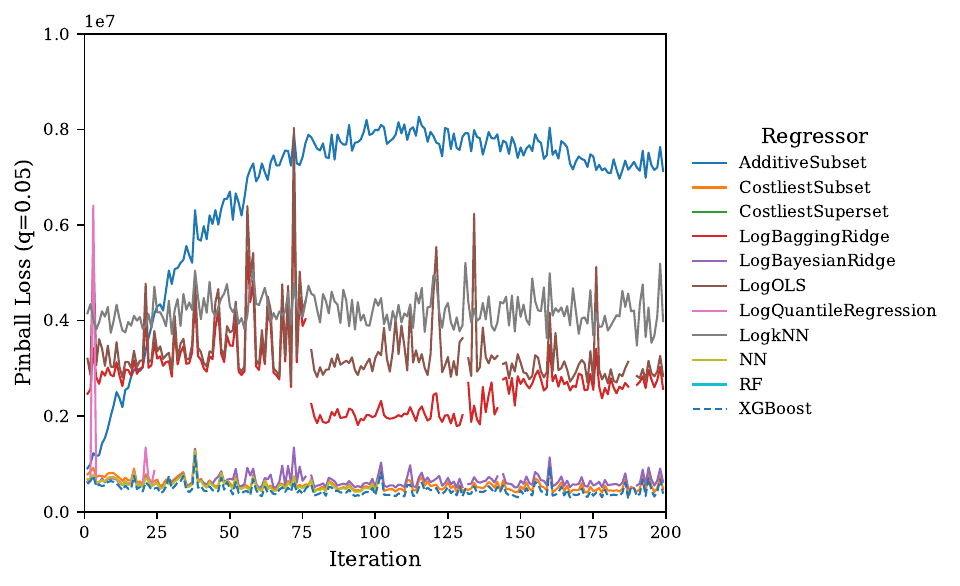}
    \caption{Pinball loss of regression models over iterations. Some models had outliers during testing, which are removed for clarity.}
    \label{fig:pinball_loss_iterations}
\end{figure}

Of the models that did complete the evaluation, several performed poorly relative to the CSH. Specifically, LogBayesianRidge, LogkNN, LogOLS, and LogBaggingRidge all exhibited higher error metrics and were unable to consistently outperform the heuristic baseline. Of the remaining models, XGBoost stands out across all evaluation metrics. Notably, XGBoost achieved a mean absolute percentage error (MAPE) 39\% lower than the next-best model in that category (LogBaggingRidge), and a pinball loss 20\% lower than the next-best model (CostliestSuperset). The Pinball loss is shown in more detail in Figure~\ref{fig:pinball_loss_iterations}. Pinball loss penalizes underpredictions and overpredictions asymetrically (19 times more with $q=0.05$), making it well suited for measuring conservatism-biased accuracy.

These results suggest that XGBoost, trained on a combination of one-hot encodings, engineered features, and heuristic outputs, provides the most accurate and conservative predictions of traffic simulation outcomes. Its ability to deliver reliable estimates under limited data and iterative retraining conditions makes it well suited for use as a surrogate model in large-scale maintenance scheduling.

\section{Conclusion}
\label{sec:conclusion}

In this paper, we developed and evaluated regression models to predict total travel time in a traffic network undergoing road renovations, providing a fast alternative to expensive traffic simulations. Our models use a range of input representations, including one-hot encoding, pairwise interactions, and engineered features derived from network properties. We benchmarked multiple models—linear, ensemble-based, probabilistic, neural, and heuristic—within an online learning framework using the canonical Sioux Falls case.

The results demonstrate that the CostliestSubset heuristic (CSH) is a competitive heuristic, with the AdditiveSubset and CostliestSuperset heuristics performing poorly. Among regression models, many performed either worse or similarly to the CSH. The notable exception was XGBoost, which achieved the best performance by a large margin. It achieved a mean absolute percentage error (MAPE) of 11\%, outperforming the next-best model by 39\%, and a pinball loss 19\% lower than the next-best models. Some models that used Pinball loss as a training measure (Neural Network, Random Forest and Quantile regression), required more training time for the online learning setting than was allowed (10 minutes per iteration). 

Our findings highlight that approximating traffic simulation outcomes is a nontrivial task. The CSH heuristic performs well when only limited data is available, making it a strong baseline. As more data becomes available, predictive accuracy improves significantly when models are trained on a well-chosen combination of engineered features and one-hot encodings. Future work could evaluate surrogate models in a true online learning setting or apply the approach to larger or more complex networks.

\bibliographystyle{splncs04}
\bibliography{references2}

\end{document}